\documentclass[10pt,twocolumn,letterpaper]{article}

\usepackage{cvpr}
\usepackage{times}
\usepackage{epsfig}
\usepackage{graphicx}
\usepackage{amsmath}
\usepackage{amssymb}
\usepackage{algorithm}
\usepackage[noend]{algpseudocode}
\usepackage{color}        
\usepackage{multirow}
\usepackage[pagebackref=true,breaklinks=true,letterpaper=true,colorlinks,bookmarks=false]{hyperref}

\cvprfinalcopy 

\def\cvprPaperID{927} 

\ifcvprfinal\fi
\begin{document}

\title{Deep Representation Learning on Long-tailed Data: A Learnable Embedding Augmentation Perspective}

\newcommand*\samethanks[1][\value{footnote}]{\footnotemark[#1]}
\author{%
	Jialun Liu$^1$\thanks{Equal contribution.}, Yifan Sun$^2$\samethanks, Hantao Hu$^3$, Chuchu Han$^4$, Zhaopeng Dou$^5$, Wenhui Li$^1$\thanks{Corresponding author.}\\
	{$^1$Jilin University}          
	{$^2$Megvii Inc.}
	{$^3$Beihang University}\\
	{$^4$Huazhong University of Science and Technology }
	{$^5$Tsinghua University}\\
	{\texttt{\small{\ jialun18@mails.jlu.edu.cn}} \hspace{0.5cm}}
	{\texttt{\small{\ peter@megvii.com}}\hspace{0.5cm}}
	{\texttt{\small{\ liwh@jlu.edu.cn}}}
}
\maketitle
\thispagestyle{empty}

\begin{abstract}
 This paper considers learning deep features from long-tailed data. 
 We observe that in the deep feature space, the head classes and the tail classes present different distribution patterns. The head classes have a relatively large spatial span, while the tail classes have significantly small spatial span, due to the lack of intra-class diversity. This uneven distribution between head and tail classes distorts the overall feature space, which compromises the discriminative ability of the learned features. Intuitively, we seek to expand the distribution of the tail classes by transferring from the head classes, so as to alleviate the distortion of the feature space. To this end, we propose to construct each feature into a `` feature cloud''. If a sample belongs to a tail class, the corresponding feature cloud will have relatively large distribution range, in compensation to its lack of diversity. It allows each tail sample to push the samples from other classes far away, recovering the intra-class diversity of tail classes. Extensive experimental evaluations on person re-identification and face recognition tasks confirm the effectiveness of our method.
\end{abstract}

\section{Introduction}
 \begin{figure*}[t!]
	\centering
	\includegraphics[width=0.9\linewidth]{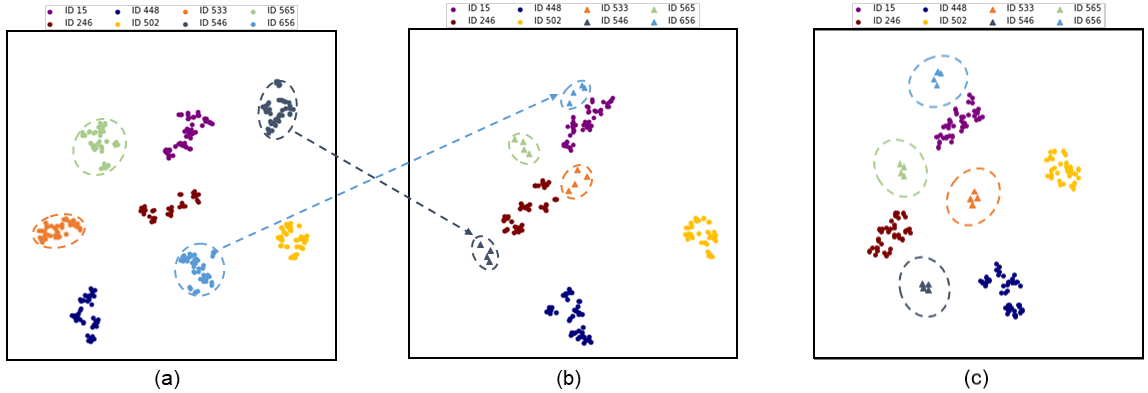}
	\caption{The visualization of features in the embedding layer with t-SNE~\cite{van2014accelerating}. (a) The visualization of features from 8 head classes (dot). With the wide region in the feature space, each class can be well distinguished. (b) We reduce the samples of some head classes so that they become tail classes (triangle). With these tail classes, the spanned feature space is narrowed, which leads to the distortion of the original feature space. So it is hard for the tail classes to be separated from other classes. (c) In training, the space is expanded for the tail class so that it is pushed away from others.
	}
	\label{fig:1}
	\vspace{-3mm}
\end{figure*}
 Large-scale datasets play a crucial role in training a model with good discriminability. However, in the real-world, large-scale datasets often exhibit extreme long-tailed distribution~\cite{everingham2010pascal,guo2016ms}. Some identities have sufficient samples, while for other massive identities, only very few samples are available. They are defined as the head classes and tail classes, respectively. With this distribution, deep neural networks have been found to perform poorly on tail classes~\cite{buda2018systematic}.
 
 The issue is clearly shown in Fig.~\ref{fig:1}. Firstly, we select eight head classes from DukeMTMC-reID dataset~\cite{zheng2017unlabeled,ristani2016performance}, and the visualization of features is shown in Fig.~\ref{fig:1} (a). It is observed that the head classes have relatively large spatial span.
 With larger inter-class distances, the head classes can be well distinguished. This observation is consistent with~\cite{yin2019feature}. Further, we reduce the samples of some head classes so they are marked as tail classes. As shown in Fig.~\ref{fig:1} (b), we observe that samples from tail class distribute narrowly in the learned feature space, due to the lack of intra-class diversity. This uneven distribution between head and tail classes distorts the overall feature space and consequentially compromises the discriminative ability of the learned features. The phenomenon indicates that when the class-imbalance exists, the feature distribution is closely related to the number of class samples. Since the tail classes with scanty training samples cannot provide sufficient intra-class diversity for learning discriminative features, they cannot be accurately distinguished from other classes. 
 
 With this insight, we propose to transfer the intra-class distribution of head classes to tail classes in the feature space. We model the distribution of angles between features and the corresponding class center, which can reflect the distribution of the intra-class features. We make a statistical analysis of the intra-class angular variance. Under a setting of person re-identification $\left \langle H20,S4 \right \rangle$, where $H$ is the number of head classes and $S$ is the number of samples per tail class, in baseline\cite{wang2018cosface}, the variations of head classes are centered at 0.463 ($\pm 0.0014$), and that of tail classes are centered at 0.288 ($\pm 0.0089$). It clearly shows that 1) tail classes have smaller variance and 2) the sample number per class is the dominating factor on the variance. Our target is to encourage the tail classes to achieve similar intra-class angular variability with the head classes in training. Specifically, we first calculate the distribution of angles between the features of head class and their corresponding class center. By averaging the angular variances of all the head classes, we obtain the overall variance of head classes. Next, we consider transferring the variance of head class to each tail class. To this end, we build a feature cloud around each tail instance in the embedding layer, and several pseudo features can be sampled with the same identities. Each instance with the corresponding feature cloud will have a relatively large distribution range, making the tail classes have a similar angular distribution with head class. Our method enforces stricter supervision on the tail classes, and thus leads to higher within-class compactness. As Figure.~\ref{fig:1} (c) shows, with the compensation of intra-class diversity during training, the tail classes are separated from other classes by a clear margin. Under the setting of person re-identification: $\left \langle H20,S4 \right \rangle$, the intra-class angular variance of tail classes turn out over even lower(than the tail classes in baseline), which is cenntered at 0.201.

 Moreover, to improve the flexibility of the method, we abandon the explicit definition of head class and tail class. Compared with some methods that divide the two classes, our approach makes the calculation entirely related to the distribution of dataset, and there is no human interference.

 We summarize the contributions of our work as follows:

\begin{itemize}
   \item We propose a learnable embedding augmentation perspective to alleviate the problem of discriminative feature learning on long-tailed data, which transfers the intra-class angular distribution learned from head classes to tail classes. 
   
   \item Extensive ablation experiments on person re-identification and face recognition demonstrate the effectiveness of the proposed method.
\end{itemize}


\section{Related Work}
 \textbf{Feature learning on imbalanced datasets.} Recent works for feature learning on imbalanced data are mainly divided into three manners: re-sampling~\cite{buda2018systematic}, re-weighting~\cite{mahajan2018exploring}, and data augmentation\cite{choi2018stargan}. The re-sampling technique includes two types: over-sampling the tail classes and under-sampling the head classes. Over-sampling manner samples the tail data repeatedly, which enables the classifier to learn tail classes better. But it may lead to over-fitting of tail classes. To reduce the risk of over-fitting, SMOTE~\cite{chawla2002smote} is proposed to generate synthetic data of the tail class. It randomly places the newly created instances between each tail class data point and its nearest neighbor. The under-sampling manner~\cite{drummond2003c4} reduces the amount of data from head classes while keeping the tail classes. But it may lose valuable information on head classes when data imbalance is extreme. The re-weighting approach assigns different weights for different classes or different samples. The traditional method re-weights classes proportionally to the inverse of their frequency of samples. Cui~\textit{et~al.}~\cite{cui2019class} improve the re-weighting by the inverse effective number of samples. Li~\textit{et~al.}~\cite{li2019gradient} propose a method which down-weights examples with either very small gradients or large gradients because examples with small gradients are well-classified and those with large gradients tend to be outliers. Recently, data augmentation methods based on Generative Adversarial Network (GAN)~\cite{choi2018stargan} are popular. \cite{yin2019feature} and~\cite{gao2018low} transfer the semantic knowledge learned from the head classes to compensate tail classes, which encourage the tail classes to have similar data distribution to the head classes. All the methods divide the classes into the head or tail class, while our method abandons the constraint.
 
 \begin{figure*}[t!]
	\centering
	\includegraphics[width=1.0\linewidth]{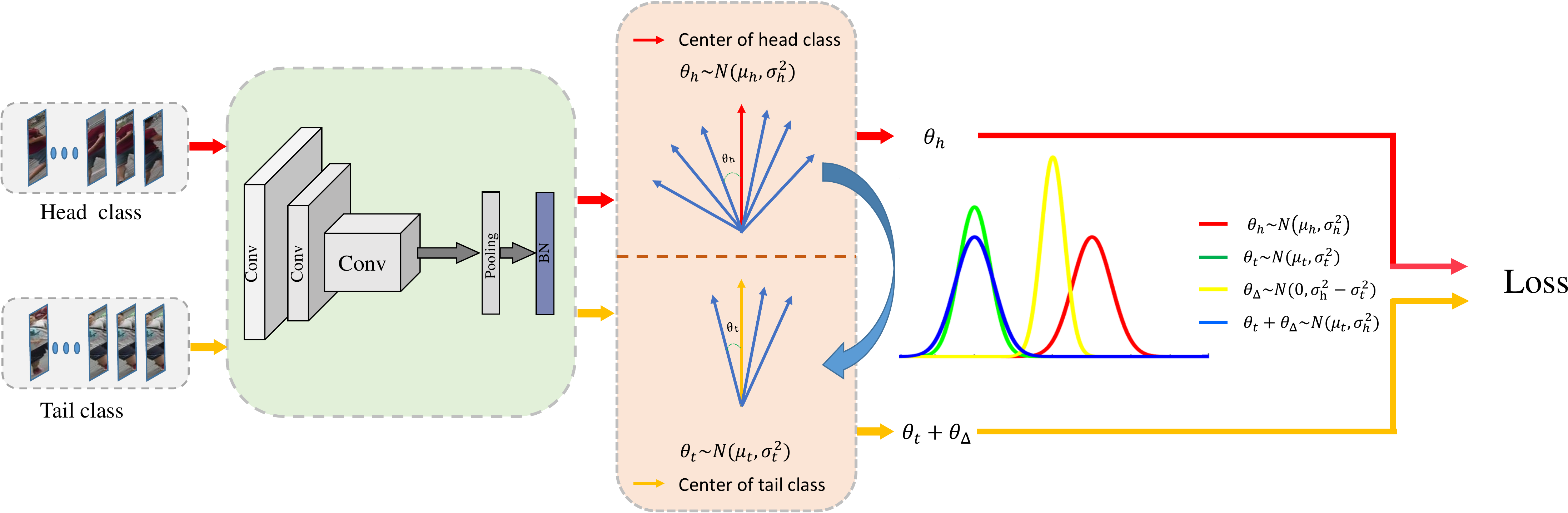}
	\caption{Overview of our proposed LEAP framework. The head data and tail data are fed into the deep network to obtain the features. We calculate the distribution of angles between the features and the class center for head class and tail class, respectively. Subsequently, we transfer the angular variance of head class ({\color{red} {red curve}}) to tail class ({\color{green} {green curve}}). In other words, based on the original distribution of tail class, we add an additional distribution ({\color{yellow} {yellow curve}}). Then we get a new distribution of tail class ({\color{blue} {blue curve}}). Finally, we use the head data and the new tail data to calculate the loss.}
	
	\label{fig:framework}
	\vspace{-3mm}
\end{figure*}
 
 \textbf{Loss function.} Loss function plays an important role in deep feature learning, and the most popular one is the Softmax loss~\cite{sun2014deep}. 
 However, it mainly considers whether the samples can be correctly classified and lacks the constraint of inter-class distance and intra-class distance. In order to improve the feature discrimination, many loss functions are proposed to enhance the cosine and angular margins between different classes. Wen~\textit{et~al.}~\cite{wen2016discriminative} design a center loss to reduce the distance between the sample and the corresponding class center. The L2-Softmax~\cite{ranjan2017l2} and NormFace~\cite{wang2017normface} add normalization to produce represented features and achieve better performance. Besides normalization, adding a margin can enhance the discrimination of features by inserting distance among samples of different classes. A-Softmax Loss~\cite{liu2017sphereface} normalizes the weights and adds multiplicative angular margins to learn more divisible angular characteristics. CosFace~\cite{wang2018cosface} adds an additive cosine margin to compress the features of the same class in a compact space, while enlarging the gap of features of different classes. ArcFace~\cite{deng2019arcface} puts an additive margin into angular space so that the loss relies on both sine and cosine dynamically to learn more angular characteristics. Our baseline is  CosFace~\cite{wang2018cosface} and ArcFace~\cite{deng2019arcface}. Although we model the intra-class angle, which is similar to them, our goal is to solve the problem of discriminative feature learning on long-tailed data.

\section{The Proposed Approach}
In this section,  A brief description of our method is given in Section \ref{sec:3.1}. We review the baseline in Section \ref{sec:3.2}. We describe the updating process of the class center and the calculation of angular distribution in Section \ref{sec:3.3}. The construction of the feature cloud for a tail instance is detailed in Section \ref{sec:3.4}. 

\subsection{Overview of Framework}\label{sec:3.1}
 The framework of our method is shown in Fig.~\ref{fig:framework}. First, the head data and tail data are fed into the deep model to extract high-dimensional features. And we consider to model the distribution of intra-class features by the distribution of angles between features and their corresponding class center. Then the center of each class is calculated, as to be detailed in Section~\ref{sec:3.3}. We build an angle memory for each class, which is used to store the angles between the features and their class center. Assuming the angles obey the Gaussian distribution, the angular distributions of head class and tail class can be denoted as $\theta_h  \sim {\rm N} (\mu_h,\sigma_h^2)$ and $\theta_t  \sim {\rm N} (\mu_t,\sigma_t^2)$, respectively. Next, we transfer the angular variance learned from the head class to every tail class. Consequently, the intra-class angular diversity of tail class is similar to the head class. Specifically, we build a feature cloud around each tail instance. An instance sampled from the feature cloud has the same identity with the tail instance. The angle between them is $\theta_\Delta$ and $\theta_\Delta  \sim {\rm N} (0,\sigma_h^2 - \sigma_t^2)$. We assume the two distribution: $\theta_t  \sim {\rm N} (\mu_t,\sigma_t^2)$ and $\theta_\Delta  \sim {\rm N} (0,\sigma_h^2 - \sigma_t^2)$ are independent of each other. By transformation, the new intra-class angular distribution of tail class is built as $\theta_t +\theta_\Delta \sim {\rm N} (\mu_t,\sigma_h^2)$ in training process. Finally, we use the original features of head classes and the reconstructed features of tail classes to calculate the loss.

\subsection{Baseline Methods}\label{sec:3.2}
 The traditional softmax loss optimizes the decision boundary between two categories, but it lacks the constraint of inter-class distance and intra-class distance. CosFace~\cite{wang2018cosface} effectively minimizes intra-class distance and maximums inter-class distance by the introducing a cosine margin to maximize the decision margin in the angular space. The loss function can be formulated as:
 \begin{equation}\label{eq:3}
 L_{1} = -\frac{1}{N}\sum_{n=1}^{N}{\log{\frac{e^{s (\cos(\theta_{{y}}) - m_c)}}{e^{s (\cos(\theta_{{y}}) - m_c)} + \sum_{j\neq y}^C{e^{s \cos(\theta_{j})}}}}},
 \vspace{-1mm}
 \end{equation}
 where $N$ and $C$ are the mini-batch size and the number of total classes, respectively. $y$ is the label of $n$-th image. We define the feature vector of $n$-th image and the weight vector of class $y$ as $f_n$ and $W_{y}$, respectively. $f_n$ and $W_{y}$ are normalized by $l_2$ normalisation and the norm of feature vector is rescaled to $s$. $\theta_{y}$ is the angle between the weight $W_{y}$ and the feature $f_n$.  $m_c$ is a hyper-parameter controlling the magnitude of the cosine margin. 

 Different from CosFace~\cite{wang2018cosface}, ArcFace~\cite{deng2019arcface} employs an additive angular margin loss, which is formulated as:
 \begin{equation}\label{eq:4}
  L_{2} = -\frac{1}{N}\sum_{n=1}^{N}{\log{\frac{e^{s (\cos(\theta_{{y}} + m_a))}}{e^{s (\cos(\theta_{{y}} + m_a))} + \sum_{j\neq y}^C{e^{s \cos(\theta_{j})}} }}},
 \vspace{-1mm}\textbf{}
 \end{equation}
 where $m_a$ is an additive angular margin penalty between feature vector $f_n$ and its corresponding $W_{y}$. It aims to enhance the intra-class compactness and inter-class distance simultaneously.
  
 In this paper, we choose CosFace~\cite{wang2018cosface} and ArcFace~\cite{deng2019arcface} as baseline. The reasons are as follows:
 \begin{itemize}
     \item They have achieved the state-of-the-art performance in the face recognition task, which can be seen as strong baselines in the community of deep feature learning.
     \item They optimize the intra-class similarity by achieving much lower intra-class angular variability. Since our method employs intra-class angles to model the intra-class feature distribution, the two loss functions can be naturally combined with our method.
 \end{itemize}

\subsection{Learning the intra-class angular distribution}\label{sec:3.3}
 The intra-class angular diversity can intuitively show the diversity of intra-class features. In this section, we study the distribution of angles between the features and their corresponding class center. $c_i$ denotes the $i$-th class center of features. $f^k_i$ is the $k$-th instance feature of class $i$. $c_i$ has the same dimension as $f^k_i$. So, we can calculate the angle between $f^k_i$ and $c_i$ as follow:

\begin{equation}\label{eq:5}
{\beta _{i,k}} = arccos(\frac{{f_i^k{c_i}}}{{||f_i^k||||{c_i}||}}),
\end{equation}

 where the $c_i$ should be updated in the training processing. Ideally, we need to take the entire training samples into account and average the features of every class in each epoch. Obviously, this approach is impractical and inefficient. Inspired by ~\cite{wen2016discriminative}, we also perform the update based on a mini-batch.
 In each mini-batch, the class center is computed by averaging the feature vectors of the corresponding class. To avoid the misleading by some mislabelled samples, we set a center learning rate $\gamma$ to update the class center. The updating method of $c_i$ is formulated as:
\begin{equation}\label{eq:6}
 {c_i^l} = {(1-{\gamma} ){c_i^l}+\gamma{c_i^{l-1}}},
\end{equation}
where $c_i^l$ is the center of class $i$ in $l$-th mini-batch. Each class center is updated by the center of current and previous mini-batch.
 
For the class $i$, we maintain an angle memory $\beta_i$ to store the angles between the features and their corresponding class center $c_i$. The size of angle memory is formulated as:
\begin{equation}\label{eq:7}
{S_i} = K_i \times P.
\end{equation}
$K_i$ is the sample number of the $i$-th class. $P$ is a hyper-parameter determining the angle memory per class. Then we calculate the mean $\mu_i$ and variance $\sigma_i^2$ of $\beta_i$. The angular distribution of the class $i$ is formulated as ${\rm N} (\mu_i,\sigma_i^2)$.

\subsection{Constructing the feature cloud for tail data}\label{sec:3.4}

 \begin{figure}[t!]
	\centering
	\includegraphics[width=1.0\linewidth]{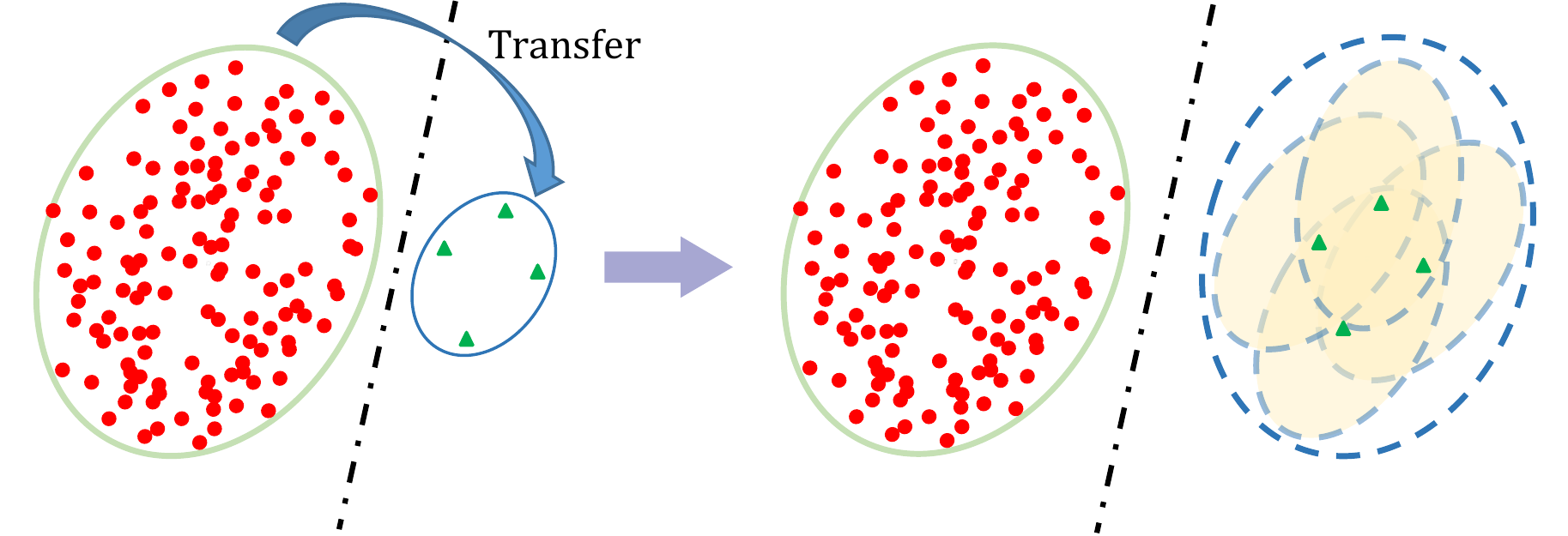}
	\caption{We transfer the intra-class angular distribution learned from the head class to the tail class. Replace each tail instance with a feature cloud.
	}
	\label{fig:transfer}
	\vspace{-3mm}
\end{figure}

 \textbf{Feature cloud}: given a specified feature $f$ of a tail class, we generate several virtual feature vectors $f'$ around it (subject to the probability distribution learned from head class), yielding the so-called ``feature cloud''.

 In this section, we elaborate the process of constructing the feature cloud for a tail instance. First, like the previous works~\cite{yin2019feature,zhong2019unequal}, we assign a label to mark the head and tail class, yielding the vanilla version of our method. On the other hand, we introduce a full version which abandons the explicit division of head and tail class. This manner is more flexible since it is only related to the distribution of the dataset.

 \textbf{Vanilla version.} We strictly divide the head class and the tail class through a threshold \textbf{$T$}. If the number of samples belonging to class $i$ is larger than \textbf{$T$}, the $i$-th class is defined as a head class. Otherwise, it is defined as a tail class.

 In the Section \ref{sec:3.3}, we have calculated the angular distribution of each class, which is assumed to lie in Gaussian distribution. By averaging the variance of all head classes, we obtain the overall variance of the head class. The mean is computed in the similar way. So the overall angular distribution of the head class is as follow:
\begin{equation}\label{eq:8}
{\mu_h} = \frac{{\sum\limits_{z = 1}^{{C_h}} {{\mu _z}} }}{{{C_h}}},\quad \quad \quad {\sigma _h^2} = \frac{{\sum\limits_{z = 1}^{{C_h}} {\sigma _z^2} }}{{{C_h}}},
\end{equation}
 where $C_{h}$ is the number of head classes. $\mu_z$ and $\sigma_z^2$ is the angular mean and variance of the $z$-th head class, respectively. $\mu_h$ and ${\sigma}_h^2$ describe the overall angular distribution of the head class. We can also obtain the class center for every tail class. The angular distribution of the $x$-th tail class is denoted as $ {\rm N} (\mu_t^x,{\sigma_t^x}^2)$. 
 
For the head classes, they include sufficient samples which show the intra-class angular diversity. In general, ${\sigma_h}$ is greater than ${\sigma_t}$, so our target is to transfer $\sigma_h^2$ to each tail class. As in Fig.~\ref{fig:transfer}, we construct a feature cloud around each feature of $x$-th tail class. By this way, the space spanned by tail class is enlarged, in training, and the real tail instances are pushed away from other classes. The angle between the feature belonging to the $x$-th tail class and a feature sampled from its corresponding feature cloud is ${\alpha_x}$, where ${\alpha_x} \sim {\rm N} (0,\sigma_h^2 - {\sigma_t^x}^2)$ and ${\alpha _x} \in \mathbb{R} {^{1 \times C}}$. In training, the feature sampled from the feature cloud shares the same identity with the real tail feature. We have assumed the two distributions: ${\rm N} (\mu_t^x,{\sigma_t^x}^2)$ and ${\rm N} (0,\sigma_h^2 - {\sigma_t^x}^2)$ are independent of each other in Section~\ref{sec:3.1}. So the original angular distribution of the $x$-th tail class is transferred from $ {\rm N} (\mu_t^x,{\sigma_t^x}^2)$ to $ {\rm N} (\mu_t^x,{\sigma_h}^2)$. 

 The new loss functions based on CosFace~\cite{wang2018cosface} and ArcFace~\cite{deng2019arcface} are defined as:

\begin{equation}\label{eq:9}
\small
L_{3} = -\frac{1}{N}\sum_{n=1}^{N}{\log{\frac{e^{s(\cos(\theta_{{y}} + \alpha_{y}) -m_c) }}{e^{s(\cos(\theta_{{y}} + \alpha_{y}) - m_c)} + \sum_{j\neq y}^{C}{e^{s \cos(\theta_{{j}} + \alpha_{y})}}}}},
\end{equation}
\begin{equation}\label{eq:10}
\small 
L_{4} = -\frac{1}{N}\sum_{n=1}^{N}{\log{\frac{e^{s(\cos(\theta_{y} + \alpha_{y} + m_a)) }}{e^{s(\cos(\theta_{{y}} + \alpha_{y} + m_a))} + \sum_{j\neq y}^C{e^{s \cos(\theta_{{j}} + \alpha_{y})}}}}},
\end{equation} 
in Eq.\ref{eq:9} and \ref{eq:10}, $\theta + \alpha$ and $\theta + \alpha + m_a$ are all clipped in the range $[0, \pi]$. $N$ and $C$ are the mini-batch size and class number, respectively. $\theta_{y}$ is the angle between the feature $f_n$ and the weight $W_{y}$ . $s$ is the scale, and $m_c$, $m_a$ are the cosine margin and the angular margin in CosFace~\cite{wang2018cosface} and ArcFace~\cite{deng2019arcface}, respectively. If $y$ is a head class, $\alpha_y$ = 0. As the training progresses, the tail class has the rich angular diversity as head class. 

Actually, we approximate the angle ($\theta'$) between the feature sampled from feature cloud and the weight. If $\alpha > 0$, we approximate $\theta'$ by the upper bound of it, and the lower bound when $\alpha \le 0$. The proof is given below.
\vspace{-4mm}
\paragraph{Proposition.} We denote a feature in the tail class as ${f}$, and $W$ is the corresponding weight vector in the full connection layer. ${f'}$ is a feature randomly sampled from the feature cloud around $f$. 
$$\small{ \langle f, W \rangle = \theta, \quad \langle f, f' \rangle = \alpha_{+}, \quad \langle W, f'\rangle=\theta'},$$
$$\small{ \|f\| = \|w\| = \|f'\| = 1, \quad 0 \le \theta+\alpha_{+} \le \pi },$$
where $\langle a,b \rangle$ represents the angle between vector $a$ and $b$, and $\|a\|$ represent the norm of vector $a$. We want to prove: $|\theta-\alpha_{+}| \le \theta' \le \theta + \alpha_{+}$.
\vspace{-4mm}
\paragraph{Proof.}Simply, we suppose that $f = [1, 0, \cdots, 0]$, then $W = [\cos{\theta}, w_2, \cdots, w_n]$. We use the Householder transformation ~\cite{householder1958unitary} to transform $W$ to $V$, where $V = [\cos{\theta}, \sin{\theta}, 0, \cdots, 0]$. Let $P = I - 2{U \cdot {U^T}}$, where $U = {W-V} / \|W-V\|$, then $f = Pf, V=PW, \hat{f'}= Pf'$. $P$ is an orthogonal transformation which preserves the inner product and norm. Therefore, we have
$$\langle f, V \rangle = \theta, \quad \langle f, \hat{f'} \rangle= \alpha_{+}, \quad \langle V, \hat{f'} \rangle = \theta'.$$
Denote $\hat{f'}=[\hat{f_1'}, \hat{f_2'}, \cdots, \hat{f_n'}]$, then
$$
\cos\alpha_{+}  = f \cdot \hat{f'} = \hat{f_1'}, \quad \hat{f_2'}^2 + \cdots  + \hat{f_n'}^2 = {\sin ^2}\alpha_{+}.
$$
We get $\hat{f_2'}\sin \theta  \in [- sin\alpha_{+} sin\theta ,sin\alpha_{+} sin\theta ]$, where $\theta  \in [0,\pi ]$. Further, we have
$$
{\cos {\theta'}} = {{\hat{f}'} \cdot V} = \cos \alpha_{+} \cos \theta   + \hat{f_2'} \sin \theta ,
$$
$$
{\cos{\theta'}} \in {[cos(\theta + \alpha_{+}), cos(\theta - \alpha_{+})]}.
$$
We get the conclusion: $|\theta-\alpha_{+}| \le \theta' \le \theta + \alpha_{+}$.

Although $\alpha \sim{N}(0, \sigma^2)$, we only need to focus on $\alpha  \in [ - \pi ,\pi ]$, since $\theta +\alpha$ is clipped in the range $[0, \pi]$.
\begin{itemize}
    \item when $0 \le \alpha \le \pi$, substituting $\alpha$ for $\alpha_{+}$, we have $|\theta-\alpha|\le \theta'\le\theta+\alpha$, in which $\theta+\alpha$ is the upper bound. 
    \item when $-\pi \le \alpha \le 0$, substituting $-\alpha$ for $\alpha_{+}$, we have $|\theta - (-\alpha)| \le \theta' \le \theta + \alpha$, which is equivalent to $\theta+\alpha \le \theta' \le \theta-\alpha$, so $\theta+\alpha$ is the lower bound.
\end{itemize}

\textbf{Full version.} The distorted feature space is well repaired by constructing a feature cloud around a tail instance. But the process in the vanilla version is inflexible. We need to set a threshold $T$ to divide the head and tail classes, artificially. The overall angular distribution in Eq.\ref{eq:8} only depends on the head classes. In the full version, the explicit definition is discarded. We have observed that the intra-class diversity is positively correlated with the number of samples, in general. Therefore, we calculate the overall variance by weighting the angular variance of each class. The weight is the number of samples in each class. The final variance is formulated as:
\vspace{-5mm}
 
\begin{equation}\label{eq:11}
\vspace{-1mm}
{\sigma}^2 = \sum\limits_{i = 1}^C {\frac{{({K_i} - 1)\sigma _i^2}}{{\sum {{\rm{(}}{{\rm{K_i}}} - 1)} }}} ,
\end{equation}
where $C$ is the number of classes, and $K_i$ is the number of samples belong to class $i$. $\sigma_i^2$ is the angular variance of the $i$-th class. A smaller $K_i$ means that the variance of the $i$-th class almost has no contribution to the final variance, so the final variance mainly depends on the classes with sufficient samples. For $i$-th class, if $\sigma _i^2 < \sigma ^2$, it means the class $i$ has poor intra-class diversity. Therefore $\alpha$ is available in Eq.\ref{eq:9} and \ref{eq:10}, and we construct the feature cloud for each instance sampled from class $i$. 

The advantage of the full version is that the calculation of feature cloud entirely depends on the distribution of the dataset. There is no human interference in the process.


\section{Experiments}
In this section, we conduct extensive experiments to confirm the effectiveness of our method. First we describe the experimental settings. Then we show the performance on person re-identification and face recognition with different long-tailed settings.
\subsection{Settings}
\textbf{Person re-identification.} Evaluations are conducted on three datasets: Market-1501~\cite{zheng2015scalable}, DukeMTMC-reID~\cite{ristani2016performance,zheng2017unlabeled} and MSMT17~\cite{wei2018person}. To study the impact of the ratio between head classes and tail classes on training a person re-identification system, we construct several long-tailed datasets based on the original dataset. We rank the classes by their number of samples. The top $150, 100, 50$ and $20$ identities are marked as the head class, respectively. The rest is treated as the tail classes, and the number of samples is reduced to $5$ each class. In this way, we form the training sets of  $\left \langle H150,S5 \right \rangle$, $\left \langle H100,S5 \right \rangle$, $\left \langle H50,S5 \right \rangle$, and $\left \langle H20,S5 \right \rangle$. For training, we choose the widely used ResNet-50~\cite{he2016deep} as the backbone. The last layer of the network is followed by a Batch Normalization layer (BN). The optimizer is Adam. The scale $s$ and $m_c$ of CosFace~\cite{wang2018cosface} are set to be $24$ and $0.2$, respectively. The scale $s$ and $m_a$ of ArcFace~\cite{wang2018cosface} are set to be $16$ and $0.2$, respectively. The learning rate of class center $\gamma$ is set to be $0.1$. For testing, the 2048-d global features after BN are used for evaluation. The cosine distance of features is computed as the similarity score. We use two evaluation metrics: Cumulative Matching Characteristic (CMC) and mean average precision(mAP) to evaluate our method.

\textbf{Face recognition.} We adopt the widely used dataset MS-Celeb-1M for training. The original MS-Celeb-1M data is known to be very noisy, so we clean the dirty face images and exclude the $79K$ identities and $1M$ images. We rank the classes through the number of samples they have. The top $5K$ and $3K$ are selected as head classes. Among the rest classes, we select the first $10K$ and $20K$ as tail classes and randomly pick $5$ images per class. In this way, we form the training set of $\left \langle H5K,T20K \right \rangle$, $\left \langle H5K,T10K \right \rangle$, $\left \langle H3K,T20K \right \rangle$ and $\left \langle H3K,T10K \right \rangle$. The face images are resized to $112 \times 112$. For training, we choose the ResNet-18~\cite{he2016deep} as our backbone. We train the model for $30$ epoch by adopting the triangular learning rate policy\cite{smith2017cyclical}, and construct feature cloud at the start of the third cycle. The scale $s$ and $m_c$ of CosFace~\cite{wang2018cosface} are set to be $64$ and $0.35$. The scale $s$ and $m_a$ of ArcFace~\cite{wang2018cosface} are set to be $64$ and $0.5$. We extract $512$-D features for model inference. For testing, we evaluate our method on LFW~\cite{huang2008labeled}, MegaFace challenge1 (MF1)~\cite{kemelmacher2016megaface} and IJB-C~\cite{maze2018iarpa}. We report our results on the Rank-1 accuracy of LFW and MF1, and different TPR@FPR of IJB-C TPR@FPR.

\vspace{-1mm}
\begin{figure*}[t!]
	\centering
	\includegraphics[width=1.0\linewidth]{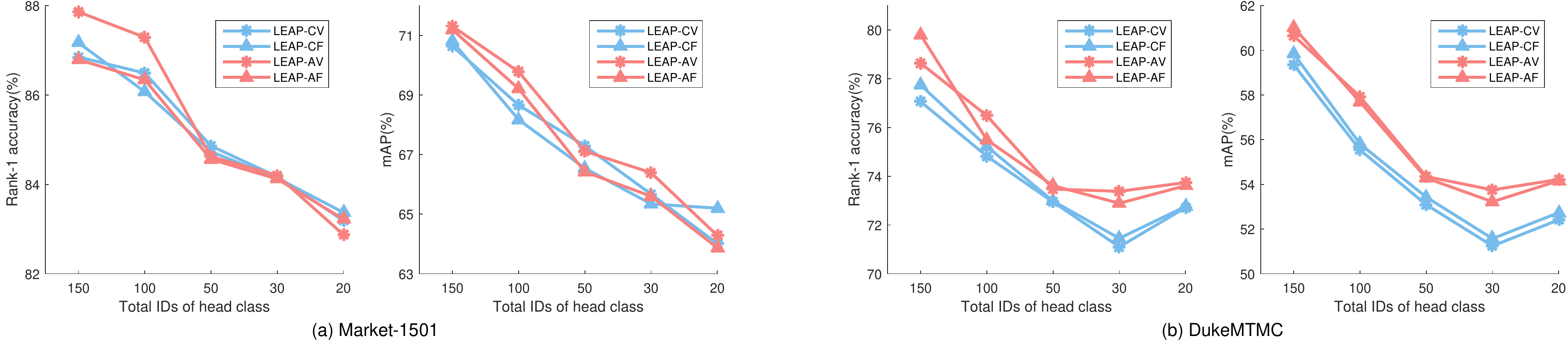}
	\vspace{-1mm}
	\caption{Comparison of vanilla version and full version on Market-1501 and DukeMTMC-reID. LEAP-CV and LEAP-AV are our vanilla version combined with CosFace and ArcFace, respectively. LEAP-CF and LEAP-AF are our full version combined with CosFace and ArcFace, respectively.}
	\label{V&F_comparison}
\end{figure*}

\begin{table}[h]
\small
\setlength{\tabcolsep}{3.5pt}
\begin{tabular}{l|cc|cc|cc}
\hline
\multirow{2}{*}{Methods} & \multicolumn{2}{c|}{Market-1501} & \multicolumn{2}{c|}{DukeMTMC} & \multicolumn{2}{c}{MSMT17} \\ \cline{2-7} 
                         & mAP         & Rank-1        & mAP          & Rank-1         & mAP         & Rank-1        \\ \hline \hline
HA-CNN~\cite{li2018harmonious}                   & 75.7        & 91.2          & 63.8         & 80.5           & -           & -             \\
PCB~\cite{sun2018beyond}                      & 77.4        & 92.3          & 66.1         & 81.8           & 40.4        & 68.2          \\
Mancs~\cite{wang2018mancs}                    & 82.3        & 93.1          & 71.8         & 84.9           & -           & -             \\
 \hline
CosFace                  & 79.5        & 92.4          & 73.0         & 85.6           & 49.2        & 75.3          \\
ArcFace                  & 81.1        & 92.5          & 73.2         & 85.8           & 50.5        & 75.5          \\ \hline
\end{tabular}
\\

\caption{Comparison with the advanced methods on the Market-1501, DukeMTMC-reID and MSMT17 datasets}
\label{basel}
\end{table}

\begin{table}[h]
\small
\setlength{\tabcolsep}{5.9pt}
\begin{tabular}{l|l|cc|cc}
\hline
\multicolumn{2}{c|}{\multirow{2}{*}{Methods}} & \multicolumn{2}{c|}{Market-1501} & \multicolumn{2}{c}{DukeMTMC} \\ \cline{3-6} 
\multicolumn{2}{c|}{}                         & mAP            & Rank-1          & mAP           & Rank-1        \\ \hline \hline
\multirow{7}{*}{GF}          & SVDNet~\cite{sun2017svdnet}         & 62.1           & 82.3            & 56.8          & 76.7          \\
                             & BraidNet~\cite{wang2018person}       & 69.5           & 83.7            & 69.5          & 76.4          \\
                             & CamStyle~\cite{zhong2018camera}            & 71.6         & 89.5            & 57.6          & 78.3         \\
                             & Advesarial~\cite{huang2018adversarially}     & 70.4           & 86.4            & 62.1          & 79.1          \\
                             & Dual~\cite{du2018interaction}           & 76.6           & 91.4            & 64.6          & 81.8          \\
                             & Mancs~\cite{wang2018mancs}          & 82.3           & 93.1            & 84.9          & 71.8          \\ 
                             & IANet~\cite{hou2019interaction}     & 83.1           & 94.4   & 73.4          & 87.1          \\
                             & DG-Net~\cite{zheng2019joint}                          & 86.0              & \textbf{94.8}    & \textbf{74.8}             & 86.6 \\ \hline
\multirow{4}{*}{PF}          & AACN~\cite{xu2018attention}           & 66.9           & 85.9            & 59.2          & 76.8          \\
                             & PSE~\cite{saquib2018pose}            & 69.0           & 87.7            & 62.0          & 79.8          \\
                             & PCB~\cite{sun2018beyond}            & 77.4           & 92.3            & 66.1          & 81.8          \\
                             & SPReID~\cite{kalayeh2018human}         & 81.3           & 92.5            & 70.9          & 84.4          \\ \hline
\multirow{2}{*}{Ours}        & LEAP-CF     & 84.2          & 94.4          & 74.2         & \textbf{87.8}        \\
                             & LEAF-AF     & 83.2         & 93.5           & 74.2        & 86.9        \\ \hline
\end{tabular}
\\
\caption{Comparison with state-of-the-art methods on Market-1501 and DukeMTMC-reID. Three groups: global features(GF), part features(PF) and ours. LEAP-CF and LEAP-AF are our full version combined with CosFace and ArcFace, respectively.}
\label{M&D}
\vspace{-2mm}
\end{table}

\subsection{Experiments on person re-identification}
 \textbf{Performance of baseline.} Table \ref{basel} reports the results of the baseline. We compare our baseline with the advanced methods. Our baseline achieves very competitive performance, which is reliable.
 
 \textbf{Comparison with state-of-the-art approaches.} 
 We compare our full version with the state-of-the-art methods on Market-1501 and DukeMTMC-reID. The comparisons are summarized in Table \ref{M&D}. It shows that our baseline has surpassed many advanced methods. And our method further improve the performance compared with baseline. Specifically, LEPA-CF achieves 94.4\% on rank-1 for Market-1501, and 87.8\% on rank-1 for DukeMTMC-reID. We further evaluate our method on a recently released large scale dataset MSMT17~\cite{wei2018person}.  The comparison is shown in Table \ref{MSMT17}. Compared with DG-Net~\cite{zheng2019joint}, our performance is very close to it. However, our method is a simple but efficient method, which does not use GAN to generate many image-level samples.
 
 \begin{table}[h]
\small
\setlength{\tabcolsep}{7pt}
\begin{tabular}{l|cccc}
\hline
Methods       & mAP  & Rank-1 & Rank-5 & Rank-10 \\ \hline \hline
GoogleNet~\cite{szegedy2015going}     & 23.0 & 47.6   & 65.0   & 71.8    \\
Pose-driven~\cite{su2017pose}   & 29.7 & 58.0   & 73.6   & 79.4    \\
Verif-Identif~\cite{zheng2018discriminatively} & 31.6 & 60.5   & 76.2   & 81.6    \\
GLAD~\cite{wei2017glad}          & 34.0 & 61.4   & 76.8   & 81.6    \\
PCB~\cite{sun2018beyond}           & 40.4 & 68.2   & 81.2   & 85.5    \\
IANet~\cite{hou2019interaction}         & 46.8 & 75.5   & 85.5   & 88.7    \\ 
DG-Net~\cite{zheng2019joint}     & \textbf{52.3} &\textbf{77.2}  &\textbf{87.4}   &\textbf{90.5} \\ \hline
LEAP-CF    & 50.8 & 76.7   & 86.9   & 90.0    \\
LEAP-AF    & 51.3 & 76.3   & 86.5   & 89.8    \\ \hline
\end{tabular}
\\
\caption{Comparison with advanced methods on the MSMT17.}
\label{MSMT17}
\vspace{-2mm}
\end{table}

 \begin{table}[h]
\small
\setlength{\tabcolsep}{5.5pt}
\begin{tabular}{cl|cc|cc}
		\hline
\multicolumn{2}{c|}{Dataset $\to$}                                                     & \multicolumn{2}{c|}{Market-1501} & \multicolumn{2}{c}{DukeMTMC} \\ \hline
\multicolumn{1}{c|}{Train $\downarrow$}                                            & Method $\downarrow$  & mAP             & Rank-1         & mAP              & Rank-1          \\ \hline \hline
\multicolumn{1}{c|}{\multirow{4}{*}{$\left \langle H150,S5 \right \rangle$}} 
& CosFace                     & 67.3                 & 86.3                       & 57.3                   & 75.6                      \\
\multicolumn{1}{c|}{}                                                  & LEAP-CV                  & \textbf{70.6}                   & \textbf{86.9}                       & \textbf{59.4}                   & \textbf{77.1}                      \\
\multicolumn{1}{c|}{}                                                  & ArcFace                     & 70.6                   & 87.3                       & 60.2                   & 77.6                     \\
\multicolumn{1}{c|}{}                                                  & LEAP-AV                  & \textbf{71.3}                   & \textbf{87.9}                       & \textbf{60.6}                   & \textbf{78.7}                      \\ \hline
\multicolumn{1}{c|}{\multirow{4}{*}{$\left \langle H100,S5 \right \rangle$}} & CosFace                     & 62.8                   & 83.3                       & 52.6                   & 70.3                      \\
\multicolumn{1}{c|}{}                                                  & LEAP-CV                  & \textbf{68.7}                   & \textbf{86.5}                       & \textbf{55.6}                  & \textbf{74.8}                      \\
\multicolumn{1}{c|}{}                                                  & ArcFace                     & 68.0                   & 86.6                       & 56.7                   & 74.8                      \\
\multicolumn{1}{c|}{}                                                  & LEAP-AV                  & \textbf{69.8}                   & \textbf{87.3}                       & \textbf{57.9}                   & \textbf{76.5}                     \\ \hline
\multicolumn{1}{c|}{\multirow{4}{*}{$\left \langle H50,S5 \right \rangle$}}  & CosFace                     & 60.5                   & 80.7                       & 48.0                   & 67.7                      \\
\multicolumn{1}{c|}{}                                                  & LEAP-CV                  & \textbf{67.3}                   & \textbf{84.9}                       & \textbf{53.1}                   & \textbf{73.0}                      \\
\multicolumn{1}{c|}{}                                                  & ArcFace                     & 64.2                  & 83.8                       & 51.1                   & 71.1                      \\
\multicolumn{1}{c|}{}                                                  & LEAP-AV                  & \textbf{67.1}                   & \textbf{84.6}  & \textbf{54.4}                   & \textbf{73.5 }                     \\ \hline
\multicolumn{1}{c|}{\multirow{4}{*}{$\left \langle H20,S5 \right \rangle$}}  & CosFace                     & 55.6                   & 78.6                       & 47.0                   & 66.0                      \\
\multicolumn{1}{c|}{}                                                  & LEAP-CV                  & \textbf{64.1}                   & \textbf{83.2}                       & \textbf{52.4}                   & \textbf{72.7}                     \\
\multicolumn{1}{c|}{}                                                  & ArcFace                     & 60.1                   & 81.1                       & 50.5                   & 69.3                      \\
\multicolumn{1}{c|}{}                                                  & LEAP-AV                  & \textbf{64.3}                   & \textbf{82.2}                       & \textbf{54.2}                   & \textbf{73.7}                      \\ \hline
\end{tabular}
\\
\caption{Controlled experiments by varying the ratio between head and tail data. $H$ is the number of head class. $S$ denotes that the number of samples per tail class. CosFace and ArcFace are baselines. LEAP-CV and LEAP-AV are vanilla versions combined with CosFace and ArcFace.}
\label{table:2}
\vspace{-2mm}
\end{table}

\textbf{Evaluation with the vanilla version.}
 We evaluate the effectiveness of the vanilla version. For comparison, we train the baseline model on the long-tailed person re-identification datasets under the supervision of CosFace~\cite{wang2018cosface} and ArcFace~\cite{deng2019arcface}. We compare our method with baseline methods. The results are shown in Table \ref{table:2}. We have the following observations. First, compared with CosFace, ArcFace has higher Rank-1 and mAP accuracy on the same long-tailed setting. For example, on Market-1501 with $\left \langle H20,S5 \right \rangle$, ArcFace achieves the Rank-1 accuracy of $81.1\%$, while the Rank-1 accuracy of CosFace is $78.6\%$. This indicates that Arcface has a stronger robustness for the long-tailed person re-identification. Second, in different long-tailed settings, the proposed LEAP method combined with CosFace and ArcFace achieves consistently better results than the baseline with significant margins.  This indicates that the LEAP is a robust method for long-tailed data distribution. Third, as the long-tailed distribution is more serious, the improvement of our method becomes even more obvious. For example, in the $\left \langle H20,S5 \right \rangle$ setting on DukeMTMC-reID, the improvement of LEAP-CV reaches $+6.7\%$ (from $66.0\%$ to $72.7\%$) in the Rank-1 accuracy. 
 
 \textbf{Comparison between vanilla version and full version.} 
 We show the results comparison of vanilla version and full version under different long-tailed settings in Figure \ref{V&F_comparison}. We observe that the full version obtains the results very close to vanilla version, and even better results in some settings. By this experiment, we justify that compared with those methods which need a label to distinguish between head class and tail class, the full version is more flexible.
 \begin{table}[h]
	\small
	\setlength{\tabcolsep}{5.5pt}
	\begin{tabular}{cl|cc|cc}
		\hline
		\multicolumn{2}{c|}{Dataset $\to$}                                                     & \multicolumn{2}{c|}{Market-1501} & \multicolumn{2}{c}{DukeMTMC} \\ \hline
		\multicolumn{1}{c|}{Train $\downarrow$}                                            & Method $\downarrow$  & mAP             & Rank-1         & mAP              & Rank-1          \\ \hline \hline
		\multicolumn{1}{c|}{\multirow{4}{*}{$\left \langle H20,S5 \right \rangle$}} & CosFace &  55.6          &  78.6         &  47.0           &  66.0          \\
		\multicolumn{1}{c|}{}                                                 & LEAP-CF &  \textbf{65.2}          &  \textbf{83.4}         &  \textbf{52.7}           &  \textbf{72.8}          \\
		\multicolumn{1}{c|}{}                                                 & ArcFace &  60.1          &  81.1         &  50.5           &  69.3          \\
		\multicolumn{1}{c|}{}                                                 & LEAP-AF &  \textbf{63.9}          &  \textbf{83.2}         &  \textbf{54.2}           &  \textbf{73.6}          \\ \hline
		\multicolumn{1}{c|}{\multirow{4}{*}{$\left \langle H20,S4 \right \rangle$}} & CosFace &  43.1          &  67.7         &  36.0           &  53.7          \\
		\multicolumn{1}{c|}{}                                                 & LEAP-CF &  \textbf{54.7}          &  \textbf{76.8}         &  \textbf{42.6}           &  \textbf{63.0}          \\
		\multicolumn{1}{c|}{}                                                 & ArcFace &  49.4          &  73.8         &  39.7           &  58.8          \\
		\multicolumn{1}{c|}{}                                                 & LEAP-AF &  \textbf{56.5}          & \textbf{77.9}         & \textbf{44.2}           & \textbf{64.4}          \\ \hline
		\multicolumn{1}{c|}{\multirow{4}{*}{$\left \langle H20,S3 \right \rangle$}} & CosFace &  31.9          &  55.5         &  25.6           &  40.8          \\
		\multicolumn{1}{c|}{}                                                 & LEAP-CF & \textbf{43.5}         & \textbf{67.2}          & \textbf{33.2}           & \textbf{51.1}         \\
		\multicolumn{1}{c|}{}                                                 & ArcFace &  36.2          &  60.1         &  28.9          & 46.7         \\
		\multicolumn{1}{c|}{}                                                 & LEAP-AF & \textbf{44.1}          & \textbf{66.1}        & \textbf{34.3}          & \textbf{53.3}         \\ \hline
	\end{tabular}
\\
\caption{ Impact analysis of different tail data for feature learning.}
\label{table:3}
\vspace{-2mm}
\end{table}

\begin{table*}[tbp]
\small
\setlength{\tabcolsep}{20pt}
\begin{center}
\begin{tabular}{c|l|c|c|ccc}
\hline
\multicolumn{2}{c|}{Test $\to$}                   & LFW    & MegaFace & \multicolumn{3}{c}{IJB-C(TPR@FPR)} \\ \hline
Train $\downarrow$                                & Method $\downarrow$        & Rank-1 & Rank-1   & 1e-3       & 1e-4       & 1e-5       \\ \hline \hline
\multirow{4}{*}{$\left \langle H5K,T10K \right \rangle$}  & CosFace    & 98.73  & 81.41    & 83.35      & 73.32      & 63.42      \\
                                                  & LEAP-CV & \textbf{98.88} & \textbf{81.78}    & \textbf{83.83}      & \textbf{73.96}      &\textbf{64.64}       \\
                                                  & ArcFace    & 98.60  & 81.08    & 82.30      & 72.45      & 62.46      \\
                                                  & LEAP-AV & \textbf{98.67}  & \textbf{81.69}    & \textbf{83.16}      & \textbf{72.97}      & \textbf{63.22}      \\ \hline
\multirow{4}{*}{$\left \langle H5K,T20K \right \rangle$}  & CosFace    & 98.87  & 82.72    & 84.77      & 76.71      & 68.19      \\
                                                  & LEAP-CV & \textbf{98.98}  & \textbf{83.16}    & \textbf{84.82}      & \textbf{77.21}      & \textbf{68.88}      \\
                                                  & ArcFace    & 98.73  & 82.76    & 84.45      & 76.22      & 66.93      \\
                                                  & LEAP-AV & \textbf{99.10}   & \textbf{83.36}    & \textbf{85.70}      & \textbf{77.77}      & \textbf{68.05}      \\ \hline
\multirow{4}{*}{$\left \langle H3K,T10K \right \rangle$}  & CosFace    & 97.65  & 72.27    & 79.08      & 68.06      & 56.52      \\
                                                  & LEAP-CV & \textbf{97.97}  & \textbf{73.19}    & \textbf{79.60}      & \textbf{69.18}     & \textbf{58.89}      \\
                                                  & ArcFace    & 97.82  & 72.45    & 78.24      & 66.99      & 55.31      \\
                                                  & LEAP-AV & \textbf{98.07}  & \textbf{73.43}    & \textbf{78.84}      & \textbf{67.82}      & \textbf{55.75}     \\ \hline
\multirow{4}{*}{$\left \langle H3K,T20K \right \rangle$}   & CosFace    & 98.02  & 74.06    & 81.21      & 71.68      & 61.03      \\
                                                  & LEAP-CV & \textbf{98.23}  & \textbf{75.18}    & \textbf{81.87}      & \textbf{72.16}      & \textbf{62.62}      \\
                                                  & ArcFace    & 98.28  & 75.24    & 81.09      & 71.36      & 61.60      \\
                                                  & LEAP-AV & \textbf{98.73}  & \textbf{76.28}    & \textbf{82.61}      & \textbf{73.21}      & \textbf{62.72}      \\ \hline
\end{tabular}
\end{center}
\caption{Face recognition results on LFW, MF1 and IJB-C are reported by varying the ratio between head and tail classes in training sets. $H$ and $T$ is the number of head class and tail class, respectively.}
\vspace{-2mm}
\label{table:5}
\end{table*}
 
 \textbf{The impact of tail data.} When the head class is reduced gradually and the tail data is increasing, the results are shown in Table \ref{table:3}, we observe the effect of tail data on performance. We gradually reduce the samples of each tail class, which results in insufficient training data, and the performance of the model drops dramatically. However, our method still makes a large margin improvement over the baseline. For example, in the $\left \langle H20,S3 \right \rangle$ setting on Market-1501, even the number of samples for each tail class is only $3$, the improvement of LEAP-CF reaches $+11.7\%$ (from $55.5\%$ to $67.2\%$) in the Rank-1 accuracy.
 
\begin{figure}[t!]
 	\centering
 	\includegraphics[width=1.0\linewidth]{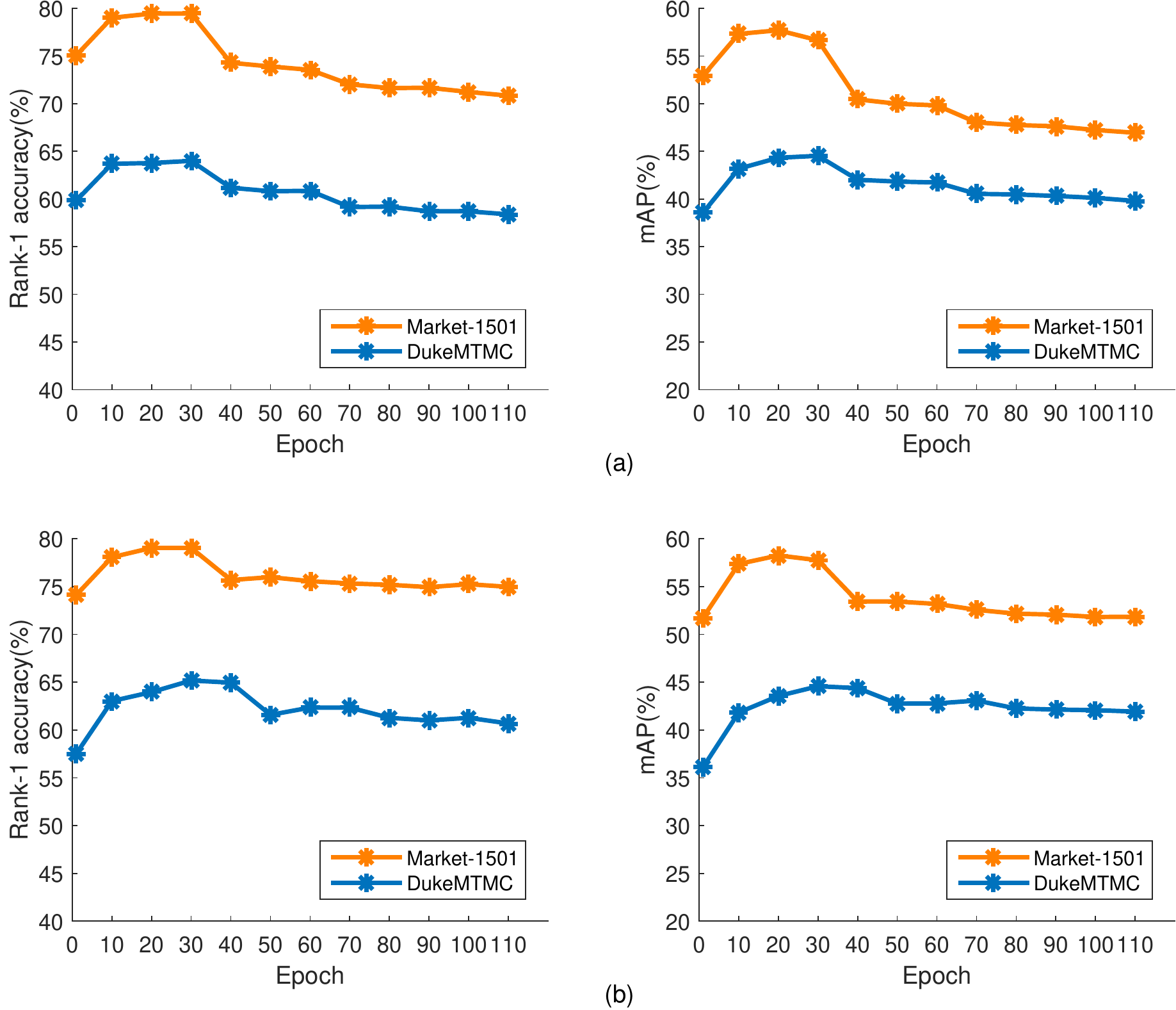}
 	\caption{Different timings of constructing the feature cloud for tail data. (a) Combined our method with CosFace~\cite{wang2018cosface}. (b) Combined our method with ArcFace~\cite{deng2019arcface} }
 	\vspace{-4mm}
 	\label{fig:timing}
 \end{figure}

 \textbf{Timing of feature cloud for tail data.} We investigate the effect of timing of constructing a feature cloud for tail data on Market-1501 and DukeMTMC-reID dataset. We take a long-tailed version: $\left \langle H20,S4 \right \rangle$ as an example. The varying curve of the results is shown in Figure \ref{fig:timing}. (a) Combined our method with CosFace~\cite{wang2018cosface}. It can be seen that, when epoch is in the range of $10$ to $30$, our results are just marginally impacted and the best results are achieved. (b) Combined our method with ArcFace~\cite{deng2019arcface}. Our results are impacted just marginally and the best results are achieved from $20$-th to $30$-th epoch. 


\subsection{Experiments on face recognition}
To further verify the observations in the person re-identification task, we perform a similar set of experiments on the face recognition task. Different from person re-identification, the dataset of face recognition has a relatively large scale. Unlike the angle memory and class center are updated frequently, this will result in a huge storage burden and computing time. In order to improve the training efficiency, we reduce the update frequency to every 5 iterations. The result is shown in Table \ref{table:5}. On LFW, our performance is improved slightly since LFW has been well solved. MF1 and IJB-C are the most challenging testing benchmark for face recognition. We report the Rank-1 accuracy of MF1 and TPR@FPR of IJB-C. Compared with the baseline, our method obtains consistency improvement. For example, in the $\left \langle H3K,T10K \right \rangle$ setting, we evaluate our method on IJB-C, the LEAP-CV improves TPR@FPR(1e-5) from 56.52\% to 58.89\%. in the $\left \langle H3K,T20K \right \rangle$ setting, we evaluate our method on MF1, the LEAP-CV improves Rank-1 accuracy from 74.06\% to 75.18\%.   

\section{Conclusions}
In this paper, we proposed a novel approach for feature learning on long-tailed data. We transfer the intra-class diversity learned from head class to tail class. We construct a feature cloud for each tail instance during training. Moreover, instead of distinguishing between head and tail classes, we propose a flexible solution which can learn the intra-class distribution adaptively. Experiments on person re-identification and face recognition consistently show that our method achieves consistent improvement. We sincerely thank Hantao Hu for his help.

{\small
	\bibliographystyle{ieee_fullname}
	\bibliography{egbib}
}
\end{document}